\documentclass[letterpaper, 10 pt, conference]{ieeeconf}  

\IEEEoverridecommandlockouts                              

\usepackage{graphicx}
\usepackage[utf8x]{inputenc}
\usepackage{comment}
\usepackage{amsmath, amssymb}
\usepackage{xcolor}
\usepackage{makecell}
\usepackage{mathtools, nccmath}
\usepackage{cite}
\usepackage{hyperref}
\usepackage{booktabs,multirow,array,caption}
\usepackage{hhline}
\usepackage{caption} 

\DeclareMathOperator{\E}{\mathbb{E}}

\usepackage{amsmath}
\DeclareMathOperator*{\argmax}{arg\,max}

\newcommand*{\rom}[1]{\expandafter\@slowromancap\romannumeral #1@}
\newcommand{\RomanNumeralCaps}[1]
{\MakeUppercase{\romannumeral #1}}

\title{\LARGE \bf
C-3PO: Cyclic-Three-Phase Optimization for Human-Robot Motion Retargeting based on Reinforcement Learning
}

\author{Taewoo Kim$^{1, 2}$ and Joo-Haeng Lee$^{2, 1, *}$
	\thanks{$^{1}$Dept. of Comp. SW and Eng., Korea University of Science and Technology, Daejeon, Rep. of Korea
		{\tt\small twkim@ust.ac.kr}}%
	\thanks{$^{2}$Human-Robot Interaction Research Group, ETRI, Daejeon, Rep. of Korea
		{\tt\small $^*$joohaeng@etri.re.kr}}%
}

\begin{document}

\setlength{\abovedisplayskip}{3pt}
\setlength{\belowdisplayskip}{3pt}

\maketitle
\thispagestyle{empty}
\pagestyle{empty}

\begin{abstract}
Motion retargeting between heterogeneous polymorphs with different sizes and kinematic configurations requires a comprehensive knowledge of (inverse) kinematics. Moreover, it is non-trivial to provide a kinematic independent general solution. In this study, we developed a cyclic three-phase optimization method based on deep reinforcement learning for human-robot motion retargeting. The motion retargeting learning is performed using refined data in a latent space by the cyclic and filtering paths of our method. In addition, the human-in-the-loop based three-phase approach provides a framework for the improvement of the motion retargeting policy by both quantitative and qualitative manners. Using the proposed C-3PO method, we were successfully able to learn the motion retargeting skill between the human skeleton and motion of the multiple robots such as NAO, Pepper, Baxter and C-3PO. 
\end{abstract}

\section{INTRODUCTION}

Humans can effortlessly imitate the motions of others with different body sizes or even animals. This is because the human has extraordinary motion retargeting skill that grasps the target’s motion attributes from visual information and connects it with their joints appropriately. The motion retargeting skill which is not difficult for humans, however, is challenging task for robots because it requires a complex algorithm for understanding motion attributes, proper mapping between source and target and handling exceptional cases. Though having some limitations, several methods have been proposed to teach the motion retargeting to robots. For example, direct joint mapping \cite{shahverdi2016simple, zuher2012recognition, lee2012full} and inverse kinematics (IK)-solver-based methods \cite{monzani2000using, mukherjee2015inverse} require expertise of robot kinematics and are difficult to generalize due to their different kinematic configurations. They also have a singular position problem \cite{craig2009introduction} and high IK calculation cost \cite{mukherjee2015inverse}. Recent learning-based approaches learn imitation skills from demonstrations that are collected from visual sensors \cite{lei2015whole, cole2007learning, zuher2012recognition, lee2012full}, motion capture (MoCap) \cite{koenemann2014real, ott2008motion, yamane2009simultaneous, kim2009stable} and virtual reality (VR) devices \cite{zhang2018deep, baek2003motion}. However, vision based sampling (e.g., human skeleton) is very noisy and unstable. MoCap and VR methods require additional cost and are not convenient to wear. Direct teaching (DT) \cite{grunwald2003programming, kushida2001human, tsumugiwa2002variable, schraft2005powermate} is intuitive to generate various robot motions because users can freely configure the robot postures by their hands. However, it has limitations in collecting a large number of demonstrations due to physical interaction with hardware and relevant teaching time.

\begin{figure}
	\begin{center}
		\includegraphics[width=\linewidth]{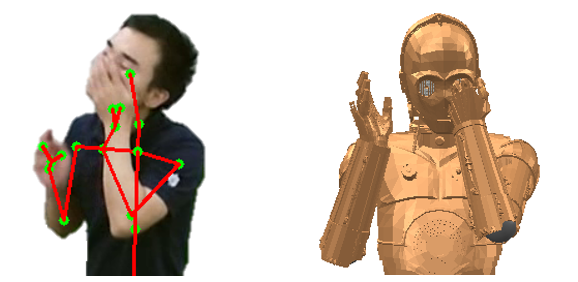}
	\end{center}
	\vspace{-1.0em}
	\caption{Motion retargeting from humans to the C-3PO robot\textsuperscript{$\dagger$} using the C-3PO algorithm.}
	\small\textsuperscript{\textsuperscript{$\dagger$} \texttt{https://www.turbosquid.com/3d-models/c-3po-star-wars-3d-obj/903731}}
	\vspace{-2.2em}
\end{figure}

In our previous work \cite{kim2019teachme}, we designed an actor-critic based simple network architecture using only the skeleton encoder and the robot motion decoder based on well-known architecture \cite{ghadirzadeh2017deep}. However, the input to the critic network was insufficient to evaluate the quality of the actor since it should be based on encoded skeleton rather than full kinematic configuration. Moreover, the encoded skeleton directly reflects the severe noise from raw input skeletons. In this paper, to overcome these limitations, we propose an advanced method to improve the three-phase framework developed on top of our previous work \cite{kim2019teachme} for learning robust human-robot motion retargeting skills. In our improved framework, the new architecture of filtering and cyclic paths are introduced to handle the noisy input and to better evaluate the actor with more abundant state information.

In reinforcement learning, the widely used temporal-difference (TD) method works effectively in the Markovian environment. If a robotic task is in the Markovian environment, the state of the robot agent should include not only the angular position but also the angular velocity to predict the next state from the current. However, low-cost motors such as Dynamixel \cite{mensink2008characterization} may not provide accurate angular velocity due to sensor errors and delays in the control system \cite{bubnov2015iterative}. Because our goal is to build a model that can be applied to such low-cost systems, we modeled our motion retargeting as a non-Markovian problem where the state of the agent has only positional information without velocity. We attempted to learn the motion retargeting policy based on the Monte-Carlo (MC) method that more effectively works in this non-Markovian environment than the TD method.

Our main contributions can be summarized as follows:

\noindent 1) We propose a novel architecture by reusing the network neglected in the previous work.

\noindent 2) Based on the newly proposed cyclic and filtering paths, we defined extended latent state and a refined reward function. This shows higher performance than our previous work.

\noindent 3) Based on a unified policy for six motion classes and an encoder-decoder network, we show that our model can sufficiently perform human-robot motion retargeting using the MC method in the non-Markovian environment.

\begin{figure}
	\begin{center}
		\includegraphics[width=\linewidth]{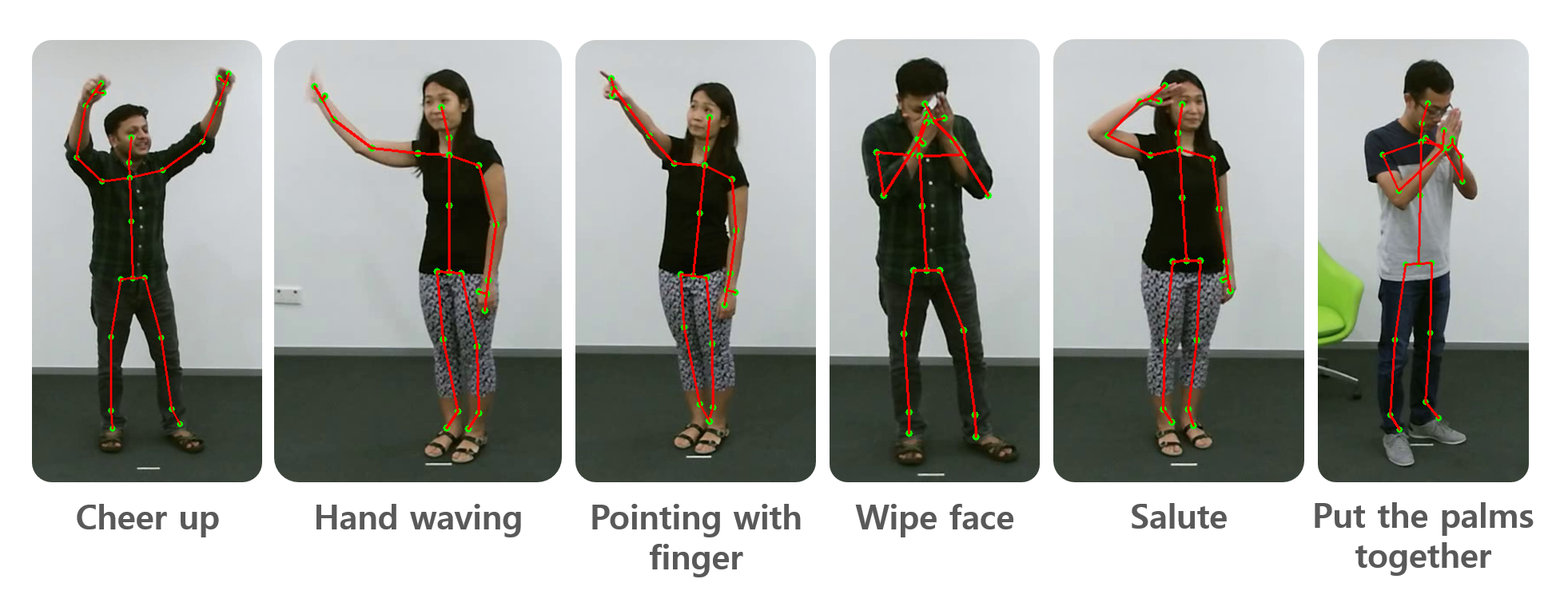}
	\end{center}
	\vspace{-1.0em}
	\caption{Our target motion classes chosen from NTU-DB.}
	\vspace{-1.5em}
\end{figure}

\section{RELATED WORK}
Motion retargeting has been attracting significant attention in many research fields including robotics and computer graphics \cite{bandera2012survey}. In this section, we review the related studies on motion retargeting and reinforcement learning.

\subsection{Motion Retargeting}
\textit{Michael} \cite{gleicher1998retargetting} proposed a method of motion retargeting on a new character with an identical kinematic structure and a different segment length using geometrically constrained optimization and a simple objective function. For online motion retargeting, \textit{Choi et al.} \cite{choi2000online} improved offline motion retargeting by space-time constraints and inverse rate control. \textit{Jean-S{\'e}bastien et al.} \cite{monzani2000using} exploited an intermediate skeleton and an IK solver for retargeting from a character's motion to a geometrically and topologically different one. Another study attempted to retarget a motion between characters with different skeleton configurations such as humans and dogs \cite{hsieh2005motion}. \textit{Ilya et al.} \cite{baran2007automatic} proposed an automatic rigging and modeling algorithm from 3D character shapes, called \textit{Pinocchio}. \textit{Chris et al.} \cite{hecker2008real} proposed a real-time motion retargeting method for highly varied user-created characters using a particle IK solver. \textit{Park et al.} \cite{park2004example} proposed an example-based motion cloning. In their work, using scattered data interpolation, the animator clones the behavior of the source example motion by specifying the key-posture between the source and the target with dynamic time-warping. They solved the time misalignment between the source and the target animation by fine-tuning the main algorithm process. 

In the robotics field, there are many studies on motion retargeting between human motions and humanoid robots. \textit{Behzad et al.} \cite{dariush2008online, dariush2009online} proposed an online motion retargeting method, which transfers human motions obtained from depth sensors to the humanoid robot ASIMO based on a constrained IK solver. \textit{Sen et al.}\cite{wang2017generative} estimated a human pose from the 3D point cloud of a depth sensor and retargeted its pose to a humanoid robot without any skeleton and joint limitations. \textit{Ko et al.} \cite{ayusawa2017motion} presented a motion retargeting method, which solves the geometric parameter identification for motion morphing and motion optimization simultaneously. With MoCap sensors, the IK-solver-based motion retargeting methods from humans to robots have been widely studied in recent years \cite{vijayan2018using, penco2018robust}. 

Although most motion retargeting studies have used IK-solver-based methods, in this study, we applied deep reinforcement learning (DRL) to motion retargeting without using any IK solvers. We also exploited the fine tuning approach for pose correction after the main learning \cite{park2004example}. 
\subsection{Reinforcement Learning}
In recent years, reinforcement learning (RL) has been used in various research areas including computer games \cite{mnih2013playing, lample2017playing, silver2016mastering}, robotics \cite{liu2018imitation} and animation \cite{peng2018deepmimic} and outperformed previous approaches. Many studies in robotics used RL for a specific task such as ball throwing \cite{ghadirzadeh2017deep}, pick \& place \cite{james2017transferring}, vision-based robotic grasping \cite{quillen2018deep}, robotic navigation \cite{faust2018prm}, and other robotic tasks in daily life \cite{levine2016end}. \textit{Peng} \cite{peng2018deepmimic} demonstrated learning skills such as locomotion, acrobatics, and martial arts on animation characters based on the reference motion and proximal policy optimization (PPO) \cite{schulman2017proximal} RL algorithm. We adopted the reference motion and the PPO algorithm with variational auto-encoder (VAE)-based \cite{kingma2013auto} network architecture \cite{ghadirzadeh2017deep} in our learning model.

\section{PRELIMINARIES}
\subsection{Deep Reinforcement Learning} 
We model motion retargeting as an infinite-horizon discounted partially observable Markov decision process (POMDP) as a tuple $\mathcal{M}=\{\mathcal{S, O, A, T}, r, \gamma, \mathbb{S}  \}$, with a state space $\mathcal{S}$, partial observation space $\mathcal{O}$, action space $\mathcal{A}$, state transition probability function $\mathcal{T}$, where $\mathcal{T}(s_{t+1}|s_t, a_t)$, reward function $r : \mathcal{S} \times \mathcal{A} \rightarrow \mathbb{R}$, discount factor $\gamma \in (0, 1]$, and initial state distribution $\mathbb{S}$. The goal of the agent is to learn a deterministic policy $\pi:\mathcal{O}\rightarrow \mathcal{A}$ that maximizes the expected discounted reward over an infinite-horizon:
\begin{equation} \label{eq1}
J=\mathbb{E_S}[R_0|\mathbb{S}]
\end{equation}
where the return is defined as follows:
\begin{equation} \label{eq2}
R_t=\sum_{i=t}^{\infty}\gamma^{i-t}r(s_i, a_i)
\end{equation}
We adopted PPO-based \cite{schulman2017proximal} actor-critic algorithm \cite{konda2000actor} to learn the policy parameters of $\omega$ for the actor and $\zeta$ for the critic network, respectively. The critic network evaluates the action-value of the policy. We define a Q-function, which describes the expected return under policy $\pi$ with parameter $\zeta$ from action $a_t$ at state $s_t$ as follows:

\begin{equation} \label{(eq3)}
\begin{split}
Q^{\pi}(s_t, a_t)&=\mathbb{E_{\pi}}[R_t|s_t, a_t] \\
				 &=\mathbb{E_{\pi}}[r(s_t, a_t) + \gamma Q^{\pi}(s_{t+1}, a_{t+1})|s_t, a_t]
\end{split}
\end{equation}
During training, the agent's experience data represented by a set of tuples $(o_t, s_t, a_t, q_t, r_t)$ are stored in a rollout memory, where $q_t=Q^{\pi}(s_t, a_t)$ and $o_t=z_t^s, s_t=(z_t^s \cup z_t^r), a_t=z_t^r$, which indicate the encoded latent representations of a skeleton $z_t^s$ and a robot posture $z_t^r$ at time $t$. The experience tuples stored in the rollout memory are then used to optimize the actor and the critic network.

\begin{figure}
	\begin{center}
		\includegraphics[width=\linewidth]{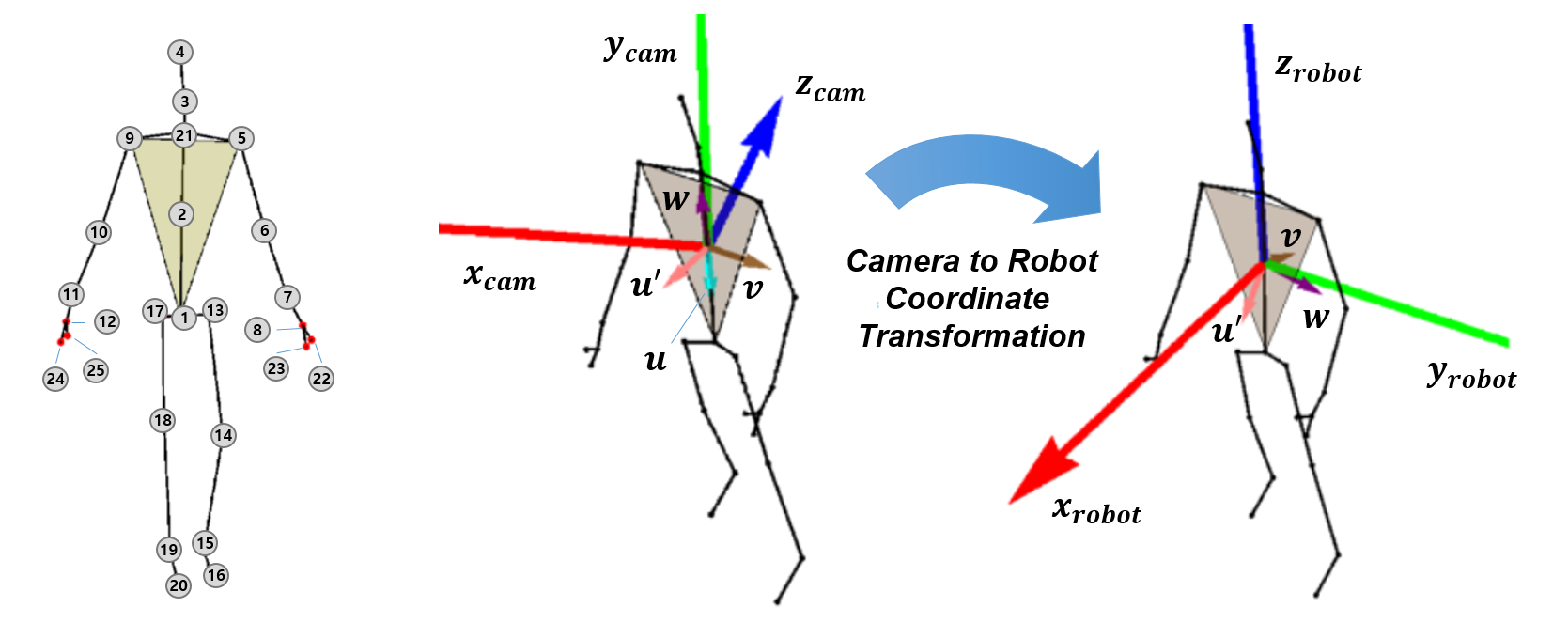}
	\end{center}
	\vspace{-1.0em}
	\caption{\textbf{[Left]} NTU-DB skeleton and each joint number. \textbf{[Right]} Transformation from camera to robot coordinates.}
	\vspace{-0.5em}
\end{figure}

\begin{table}[h]
	\vspace{-0.5em}
	\caption{NTU-DB Data Refinement Statistics.}
	\label{table_example}
	\vspace{-0.5em}
	\begin{center}
		\begin{tabular}{m{2.6cm}||m{1.8cm}|m{1.0cm}|m{1.6cm}}
			\hline
			Class Name
			& \makecell{Refined Scene\\(Filtered / Total)}  
			& \makecell{Use\\Rate}
			& \makecell{Total\\No. of Frames} \\
			\hline
			
			Cheer up 
			& \centering 533 / 948
			& \centering 56.2\% 
			& \centering 37,613 \tabularnewline
			\hline
			
			Hand waving 
			& \centering 522 / 948
			& \centering 55.0\% 
			& \centering 37,228 \tabularnewline
			\hline
			
			Pointing with finger 
			& \centering 500 / 948
			& \centering 52.7\% 
			& \centering 28,296 \tabularnewline
			\hline			
			
			Wipe face 
			& \centering 389 / 948
			& \centering 41.0\% 
			& \centering 42,172 \tabularnewline
			\hline			
			
			Salute 
			& \centering 508 / 948
			& \centering 53.5\% 
			& \centering 29,258 \tabularnewline
			\hline			
			
			Put the palms together 
			& \centering 493 / 948
			& \centering 52.0\% 
			& \centering 27,994 \tabularnewline
			\hline			
		\end{tabular}
	\end{center}
	\vspace{-3.0em}
\end{table}

\subsection{Source Dataset}
For learning human-robot motion retargeting skill, we utilized the public human motion dataset NTU-DB \cite{shahroudy2016ntu}. From initially chosen 12 motion classes among a total of 60, 6 classes such as \texttt{shake head} were ruled out because it is impossible to recognize the motion using only the skeleton. The final six motion classes are \texttt{\{cheer up, hand waving, pointing with finger, wipe face, salute, and put the palms together\}} (Fig. 2). We also excluded the data with severe noise and used 90\% of the data for training and the remaining 10\% for evaluation (Table \RomanNumeralCaps{1}). The NTU-DB data manipulation code can be found in our repository: \url{https://github.com/gd-goblin/NTU_DB_Data_Loader}.

\subsection{Data Pre-Processing}
The skeleton data of the NTU-DB are given in camera coordinates while the robot data are given based on its torso coordinates. Because the reward in phase 2 is calculated using direction vector similarities, the coordinate alignment process between the skeleton and the robot is essential. For proper alignment, we made following assumptions.

\noindent At least within the selected motion classes:
\begin{itemize}
	\item No bending posture at the waist exists.
	\item Therefore, shoulder, torso, and pelvis center joint in the skeleton are coplanar.
	\item The vector from the left shoulder joint to the right is always parallel to the ground.
\end{itemize}
Based on these assumptions, we performed coordinate alignment in two steps: 1) normalization with respect to (w.r.t) the skeleton torso frame, and 2) rotation w.r.t the robot basis frame. In the first step, each skeleton joint position $x{_i^s}=\{x, y, z\}$ is normalized by subtracting the torso position for all skeleton joints $x{_i^{s'}}=x{_i^s}-x{_{torso}^s} \ \forall i \in U$, where $U=\{1, 2, \cdots, 25\}$. For the second step, we first need to make an identical local coordinate to the robot torso frame. To do this, we get a vector $u$ by $u=x{^{s'}_{c.pelvis}}-x{^{s'}_{torso}}$, each of which corresponds to joint number 1 and 2 respectively in Fig. 3, and $v=x{^{s'}_{RShoulder}}-x{^{s'}_{LShoulder}}$, which corresponds to the 5 and 9. We can then calculate the anterior axis by $u'=u \times v$ and obtain the cranial vector by $w=u' \times v$. From the normalized local coordinate frame, we create a direction cosine matrix (DCM) and transform the skeleton in camera coordinate using the DCM transpose, which is identical to the robot basis frame matrix $I$:

\begin{gather}
\textrm{DCM}=
\begin{bmatrix} 
I_{xx} & I_{xy} & I_{xz} \\ 
I_{yx} & I_{yy} & I_{yz} \\
I_{zx} & I_{zy} & I_{zz}
\end{bmatrix}
\begin{bmatrix} 
u'_x & v_x & w_x \\ 
u'_y & v_y & w_y \\ 
u'_z & v_z & w_z 
\end{bmatrix}
\end{gather}

\begin{equation} \label{(eq5)}
x{^{s''}_i}=\textrm{DCM}^T x{^{s'}_i} \quad \forall i \in U
\end{equation}
\noindent where $x{^{s'}_i}$ and $x{^{s''}_i}$ are normalized joint positions and transformed positions to the robot coordinates, respectively.

\begin{figure*}
	\begin{center}
		\includegraphics[width=1.0\linewidth]{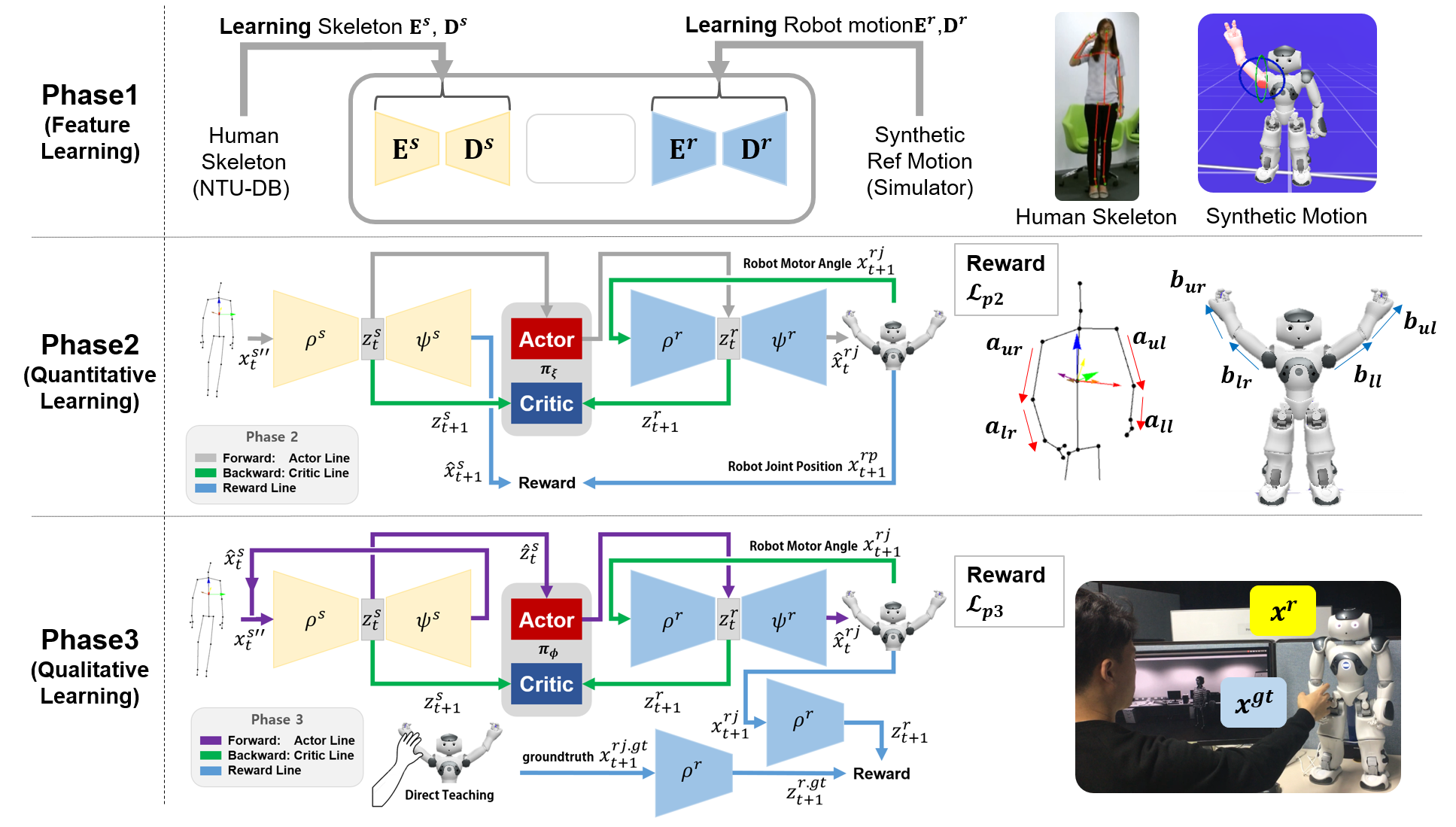}
	\end{center}
	\vspace{-1.0em}
	\caption{Cyclic-three-phase optimization framework for human-robot motion retargeting. In phase 1, latent manifold for the skeleton and the robot motion are trained using the NTU-DB and the robot reference motion. Quantitative learning is performed using a simulator and a reward function in phase 2. The policy is optimized by DT-based fine-tuning in phase 3.}
	\vspace{-1.5em}
\end{figure*}

\section{METHOD}
In this section, we describe the details of our advanced three-phase framework including filtering and cyclic paths. Also, the $n$-step MC is introduced with its formulation.

\subsection{Problem Formulation}
The skeleton generation function $f^s$ takes an image of human posture $D_t$ at time $t$ and generate a skeleton vector $x{_t^s}=f^s(D_t)$ corresponding to the input human posture, where the raw skeleton data contain x,y,z positions for all joints $x{_t^s}=\{x_1, y_1, z_1, \cdots , x_{25}, y_{25}, z_{25}\}$. The skeleton encoder $\rho^s$ then takes a transformed skeleton $x{_t^{s''}}$ (Eq.(5)) from the raw skeleton data and generates a seven-dimensional latent representation $z{_t^s}=\rho^s(x{_t^{s''}})$. The skeleton latent vector $z{_t^s}$ can be decoded by the skeleton decoder $\psi^s$ as $\hat{x}{_t^s}=\psi^s(z{_t^s})$ for later use in skeleton reconstruction and latent representation learning. Similarly, robot motion $x{_t^{rj}} = \{\theta_1, \theta_2, \cdots , \theta_{14}\}$ defined by joint angles (rad) at time $t$ is encoded by the robot motion encoder $\rho^r$ as $z{_t^r} = \rho^r (x{_t^{rj}})$. This latent vector of the robot motion $z{_t^r}$ can also be decoded by the robot motion decoder $\psi^r$ as $\hat{x}{^{rj}_t}=\psi^r(z^r_t)$ for future use in robot motion reconstruction and latent representation learning. Our mapping policy $\pi^\omega$ performs motion retargeting by mapping between the latent representations of the skeleton $z{^s_t}$ and the robot motion $z{^r_t}$ as $z{_t^r} = \pi^\omega (z{_t^s})$. 

\subsection{Phase 1: Learning Latent Manifold}
In the first phase, we learn the latent manifold of the skeleton and the robot motion using VAE \cite{kingma2013auto} (Fig. 4). The skeleton encoder $\rho^s$ consists of four fully connected (FC) layers including 512, 256, 128, 64 with ReLU, and encodes the transformed skeleton $x{^{s''}_t}$ in the seven-dimensional latent vector $z{^s_t}$. The skeleton decoder $\psi^s$ has an identical structure to the encoder but in reverse order. We created a unified skeleton encoder-decoder by learning from all six motion class data at once. As described in Table \RomanNumeralCaps{1}, the training data are randomly selected from 90\% of the refined NTU-DB.

In order to learn the robot motion encoder-decoder, we need to sample a set of reference robot motion trajectories for each class. We generated a small set of reference motion trajectories $\tau_i$ for all classes using V-REP \cite{rohmer2013v} and Choregraphe \cite{pot2009choregraphe}, where $\tau_i = \{x{_t^{rj}}, x{_{t+1}^{rj}}, \cdots , x{_{t+T}^{rj}}\} \; \textrm{for} \; \forall i$, and $i \in H=\{\texttt{cheer up}, \cdots \}$. The reference motion generation took a few minutes per class on average. Based on the reference motion, augmented training dataset were generated by adding uniform noise to $\tau_i$ iteratively as $x{_t^{rj'}}= x{_t^{rj}} + \epsilon$, where $\epsilon=[-0.05,0.05]$. We augmented our reference motion trajectories up to 20k frames per class and combined them to learn a unified robot motion encoder-decoder from a total of 120k augmented datasets. The robot motion network consists of three FC layers with Tanh including 256, 128, 64 for the encoder, 7 for latent representation, and identical but in reverse order for the decoder. MSE loss is used for both the skeleton and robot motion networks with learning rate=$1.0 \times 10^{-4}$, weight decay=$1.0 \times 10^{-6}$ and batch size=128.

\begin{figure*}
	\begin{center}
		\includegraphics[width=1.0\linewidth]{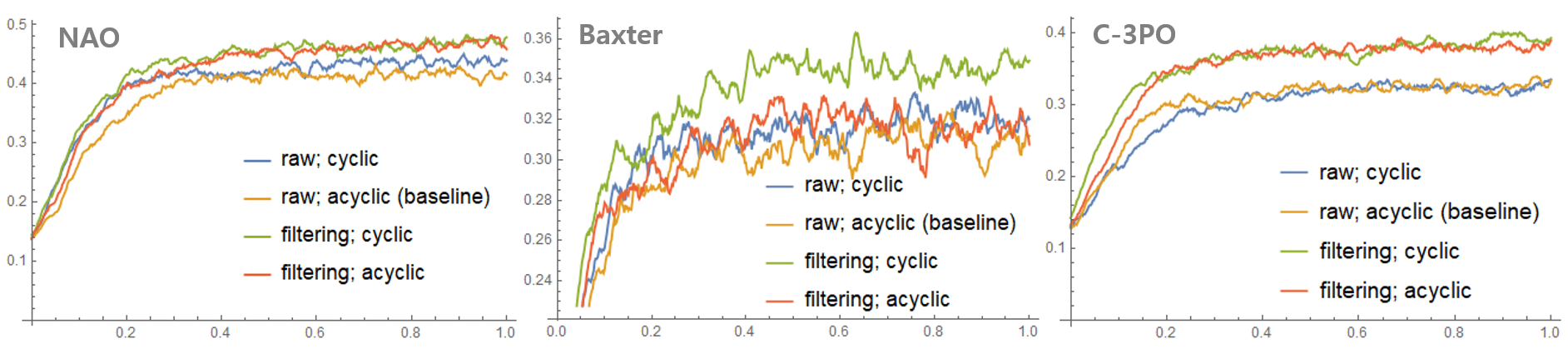}
	\end{center}
	\vspace{-1.0em}
	\caption{Training results on the NAO, Baxter, and C-3PO robot. Policies using filtering and cyclic show the best performance.}
	\vspace{-1.5em}
\end{figure*}

\subsection{Phase 2: Learning Mapping Function}
In the second phase, we learn mapping policy $\pi_\xi$ for proper motion retargeting based on a simulator and a reward function. In the forward step represented by the gray line in the second row of Fig.4, $\rho^s$ encodes a raw skeleton to a latent vector at time $t$, where $z{^s_t}=\rho^s(x{^{s''}_t})$. The actor then performs mapping to generate a robot motion latent vector $z{^r_t}=\pi{_\xi^\omega}(z{^s_t})$, and the decoded vector $\hat{x}{^{rj}_t}=\psi^r(z{^r_t})$ is transferred to the robot in the simulator. After processing one time step ($dt$=50ms), the simulator outputs the next states $x{^{rj}_{t+1}}$ with $x{^{rp}_{t+1}}$, which contains relative x,y,z positions of the robot arm w.r.t the torso frame $x{_t^{rp}}=\{x_{LS}, y_{LS}, z_{LS}, x_{LE}, y_{LE}, z_{LE}, x_{LW}, y_{LW}, z_{LW}, x_{RS}, y_{RS},\\ z_{RS}, x_{RE}, y_{RE}, z_{RE}, x_{RW}, y_{RW}, z_{RW}\}$; the subscripts represent the (left and right) shoulders, elbows and wrist joints respectively. This position vector is used in the following reward function:

\begin{equation} \label{(eq6)}
\delta_i=\arccos\bigg({a_i \cdot b_i \over ||a_i|| ||b_i||}\bigg), \ i \in \textrm{U}={\{ur, lr, ul, ll\}}
\end{equation}
\begin{equation} \label{(eq7)}
\mathcal{L}_{p2}={1 \over n} \sum_{i\in S} {\exp(-2.0 \cdot \delta_i)} 
\end{equation}
\noindent where Eq.(6) describes the reward based on the similarities in the arm vector between the skeleton and the robot. Vectors $a_i$ and $b_i$ represent the direction vector of the upper and lower left and right arms for both the skeleton and the robot (see the right side of the second row in Fig. 4). These vectors can be obtained by taking the vector difference; e.g., the upper right robot arm vector is given by $b_{ur} = (x_{RE}-x_{RS}, y_{RE}-y_{RS}, z_{RE}-z_{RS})$. Even though we ruled out the skeletons with severe noise, there still remain noisy data in our dataset. Thus, in case of the skeleton, we calculated the arm vectors from the reconstructed skeleton $\hat{x}{_t^s}=\psi^s \circ \rho^s(x{_t^{s''}})$ for denoising \cite{kingma2013auto} in the reward calculation. The cosine similarity-based reward $\delta_i$ is then normalized by multiplying the error amplitude constant (-2.0) and the exponential function, where $\mathcal{L}_{p2} \in [0, 1]$. In phase 2, there is a cyclic structure for learning the critic network based on the latent representation. The joint angle in the next step $x{^{rj}_{t+1}}$ is encoded again and combined with the next latent vector of the skeleton as $s_{t+1}=z{_{t+1}^s} \cup \rho^r(x{_{t+1}^{rj}})$. The critic network $\pi^\zeta$ evaluates the action-value of the agent based on this full state, which contains the information of the skeleton and the robot motion at time $t$. In section \RomanNumeralCaps{5}, we demonstrated the comparative analysis results that the cyclic architecture can improve the motion retargeting performance. The phase 2 objective function can be defined as:
\begin{align} \label{(eq8)}
\xi^*= \argmax_\xi \E_\pi[\mathcal{L}_{p2}(\hat{x}^s, x^{rp})|\mathbb{S}]
\end{align}
\noindent where $\hat{x}^s=\psi^s \circ \rho^s(x{^{s''}})$ and $x^{rp}$ is obtained by applying $\hat{x}^{rj} = \psi^r \circ \pi^\omega \circ \rho^s(x{^{s''}})$ to the simulator. The goal of phase 2 is to find the optimal policy parameters $\xi^*=\{ \omega^*, \zeta^*\}$ that maximizes the expected reward $\mathcal{L}_{p2}$. Our unified policy network consists of three 512 FC layers with the ReLU.

\subsection{Phase 3: Policy Optimization by Fine Tuning}
Even though the policy performs the mapping between the latent manifolds that are learned by the reference motion in phase 1, false retargeting can possibly occur because the reward of phase 2 does not consider the posture of the head or the wrist. In the last phase, we attempted to correct this false retargeting using DT-based fine tuning. First, we collected a ground truth dataset of 512 frames per class for about ten minutes using DT. The ground truth consists of a set of transformed skeleton frames $\boldsymbol{x{^s_{gt}}}=\{x{^{s''}_t}, x{^{s''}_{t+1}},\cdots\}$, corresponding robot joint angles $\boldsymbol{x{^{rj}_{gt}}}=\{x{^{rj}_t}, x{^{rj}_{t+1}}, \cdots\}$ and robot joint positions $\boldsymbol{x{^{rp}_{gt}}}=\{x{^{rp}_t}, x{^{rp}_{t+1}}, \cdots\}$. Because the reward of phase 2 was calculated using the reconstructed skeleton $\hat{x}{^s_t}$, we constructed a cyclic structure that encoded the reconstructed skeleton $\hat{z}{^s_t}=\rho^s \circ \psi^s \circ \rho^s(x{^{s''}_t})$ (see phase 3 in Fig. 4). In the forward pass, the actor retargets the latent vector of a ground truth skeleton to generate a robot motion prediction. After one step simulation, the observed next robot state and the corresponding ground truth are encoded to calculate the reward function of phase 3:
\begin{align} \label{(eq9)}
e=||\rho^r(x{^{rj}})-\rho^r(x{^{rj.gt}})||_2 + \mathcal{N}(\mu, \sigma^2)
\end{align}
\begin{align} \label{(eq10)}
\mathcal{L}_{p3}=\exp(-1.0 \cdot e)
\end{align}
\noindent where $e$ is calculated using the $\operatorname{\ell_2-norm}$ between the robot motion prediction and the corresponding ground truth in the latent space with the human teaching error ($\mu$=0, $\sigma^2$=$1.0 \times 10^{-3}$) and then normalized between 0 and 1. The following equation shows the objective function of phase 3 to determine the optimal parameter $\phi^*=\{ \omega^*, \zeta^*\}$.
\begin{equation} \label{(eq11)}
\phi^*= \argmax_\phi \E_\pi[\mathcal{L}_{p3}(\hat{z}^r, z{^{r.gt}})|\mathbb{S}(\phi)=\mathbb{S}(\xi^*)]
\end{equation}

\subsection{$n$-Step Monte-Carlo Learning}
In general, the MC method has unbiased, high variance estimates, while the TD has biased and low variance estimates. This is because MC empirically updates the policy with the actual return, whereas the TD estimates the expected rewards by inference using bootstrapping \cite{sutton2018reinforcement}. MC usually works in episodic environments; however, it can be applied to our motion retargeting because we modeled our problem as a non-episodic task and the reward can be obtained at each time frame. Owing to the continuing and every-reward environment, we can apply the n-step MC to our problem:
\begin{align}
Q^{\pi}(s_t, a_t)&=\mathbb{E_{\pi}}[G_t|s_t, a_t] \label{(eq12)} \\
				 &=\mathbb{E_{\pi}}[R_{t+1} + \gamma R_{t+2} + \cdots + , \gamma^{T-1}R_T|s_t, a_t] \nonumber
\end{align}
\noindent where $T$ represents the number of steps in n-step MC. We present the comparative results on the n-step MC and TD (Eq.(3)) method in the next section. 

\section{EXPERIMENTS}
\noindent \textbf{Intuitive Motion Retargeting.} Our C-3PO algorithm can be applied to various robots with different kinematics and sizes. This method is more intuitive than the other methods such as direct joint mapping or IK-solver-based methods because it does not require knowledge about mathematical modeling of kinematics. Through this method, we can learn the motion retargeting skill by manually appointing major joints (e.g., shoulder) and generating simple reference motion. We were successfully able to teach motion retargeting skills to the NAO, Pepper, Baxter and C-3PO robots (Fig.6).

\noindent \textbf{Learning Details.} We used the learning rates of $1.0 \times 10^{-4}$ for the actor and $2.0 \times 10^{-4}$ for the critic. The rest of the hyper-parameters were set as follows: rollout steps=2048, PPO epoch=5, mini batch size=32, $\gamma$=0.98, entropy coefficient=$5.0 \times 10^{-3}$. V-REP and Choregraphe run at 20Hz. Each unified policy learning for 1 million frames takes about 8 h on i7-8700K and Titan Xp.

\begin{figure}
	\begin{center}
		\includegraphics[width=\linewidth]{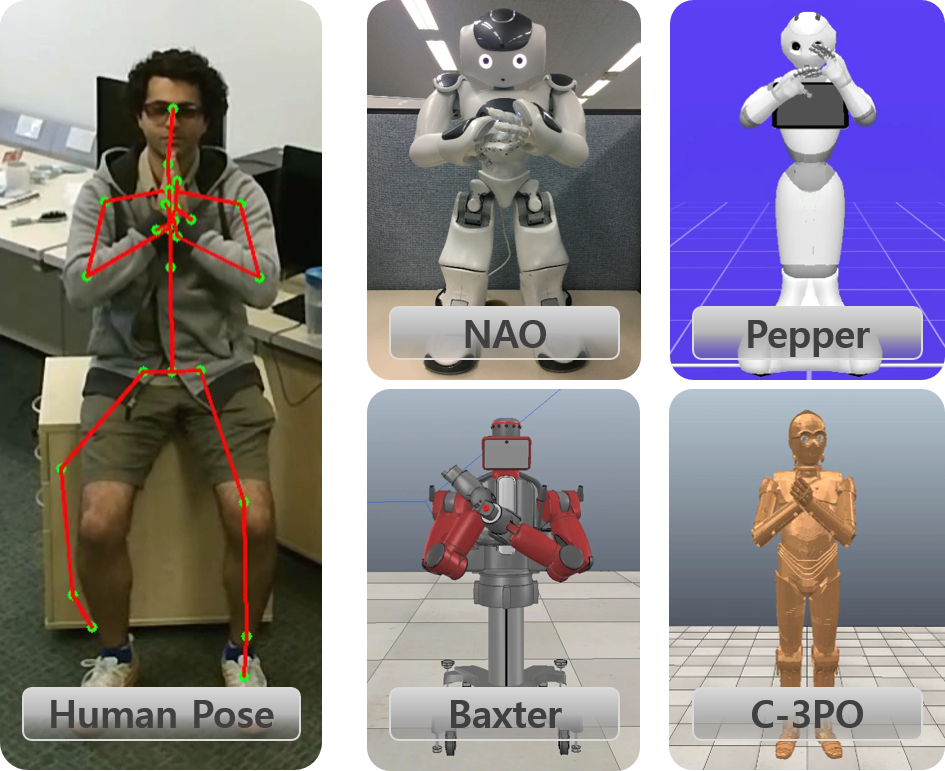}
	\end{center}
	\vspace{-1.0em}
	\caption{Motion Retargeting result using our C-3PO algorithm. The Pepper was controlled using the NAO's policy because they have identical kinematic configurations.}
	\vspace{-0.5em}
\end{figure}

\subsection{Ablation Study on Network Architecture}
To verify the performance in terms of network architecture, we evaluated them by combining the raw skeleton $x_s$, the filtered skeleton $\hat{x}_s$, and the cyclic $s_t=z{^s_t} \cup z{^r_t}$ and acyclic $s_t=z{^s_t}$ path-based reward function calculations. Fig. 5 represents the training results in these four cases. Due to the effects of noise filtering, policies using the filtering path show far better performance than the raw skeleton method. The cyclic path is also shown to assist the policy to output better action. This ablation study shows that the proposed method is effective in improving the latent space-based motion retargeting task in various types of robots with different kinematic configurations and sizes.

\begin{table}[h]
	\caption{Performance Comparison Result based on average reward and standard deviation between TD and n-step MC Methods during 5k frames.}
	\vspace{-1.0em}
	\begin{center}
		\begin{tabular}{m{0.8cm}||m{1.75cm}|m{1.28cm}|m{1.28cm}|m{1.28cm}}
			\hline
			\multirow{2}*{ } & \centering TD & \multicolumn{3}{c}{MC: filtering; cyclic} \\
			\hhline{~~---}
			& \centering filtering; cyclic
			& \centering 1-step 
			& \centering 3-steps 
			& \centering 5-steps \tabularnewline
			\hline
			
			\centering NAO
			& \makecell{$\overline{r}:0.4463$ \\ $\sigma:\textbf{0.1578}$} 
			& \makecell{$\overline{r}:0.5884$ \\ $\sigma:0.1722$}
			& \makecell{$\overline{r}:\textbf{0.5911}$ \\ $\sigma:0.1725$}
			& \makecell{$\overline{r}:0.5841$ \\ $\sigma:0.1716$} \tabularnewline
			\hline
			
			Baxter 
			& \makecell{$\overline{r}:0.2797$ \\ $\sigma:\textbf{0.1227}$} 
			& \makecell{$\overline{r}:\textbf{0.4574}$ \\ $\sigma:0.1633$} 
			& \makecell{$\overline{r}:0.4389$ \\ $\sigma:0.1523$}
			& \makecell{$\overline{r}:0.4535$ \\ $\sigma:0.1647$} \tabularnewline
			\hline
			
			C-3PO 
			& \makecell{$\overline{r}:0.4207$ \\ $\sigma:0.1543$} 
			& \makecell{$\overline{r}:0.4818$ \\ $\sigma:0.1692$} 
			& \makecell{$\overline{r}:\textbf{0.4969}$ \\ $\sigma:0.1769$}
			& \makecell{$\overline{r}:0.4609$ \\ $\sigma:\textbf{0.1532}$} \tabularnewline
			\hline			
		\end{tabular}
	\end{center}
	\vspace{-2.5em}
\end{table}

\subsection{Temporal Difference and n-step Monte-Carlo Learning}
We evaluated the performance of TD and MC w.r.t the number of steps during 5k frames. As shown in section \RomanNumeralCaps{5}.\textit{A}, because the policy using the filtering and the cyclic paths showed the best performance, we only considered that poliy in the TD method, and the n-step of MC was set to 1, 3, and 5. The experimental results evaluated by the average mean $\overline{r}$ and the standard deviation. $\sigma$ in Table \RomanNumeralCaps{2} suggest that MC ourperforms TD method in the non-Markovian motion retargeting problem. In MC, step-3 outperforms the others, but the overall performance is similar, and there is no dramatic performance improvement in more than three steps.
\begin{figure}
	\begin{center}
		\includegraphics[width=\linewidth]{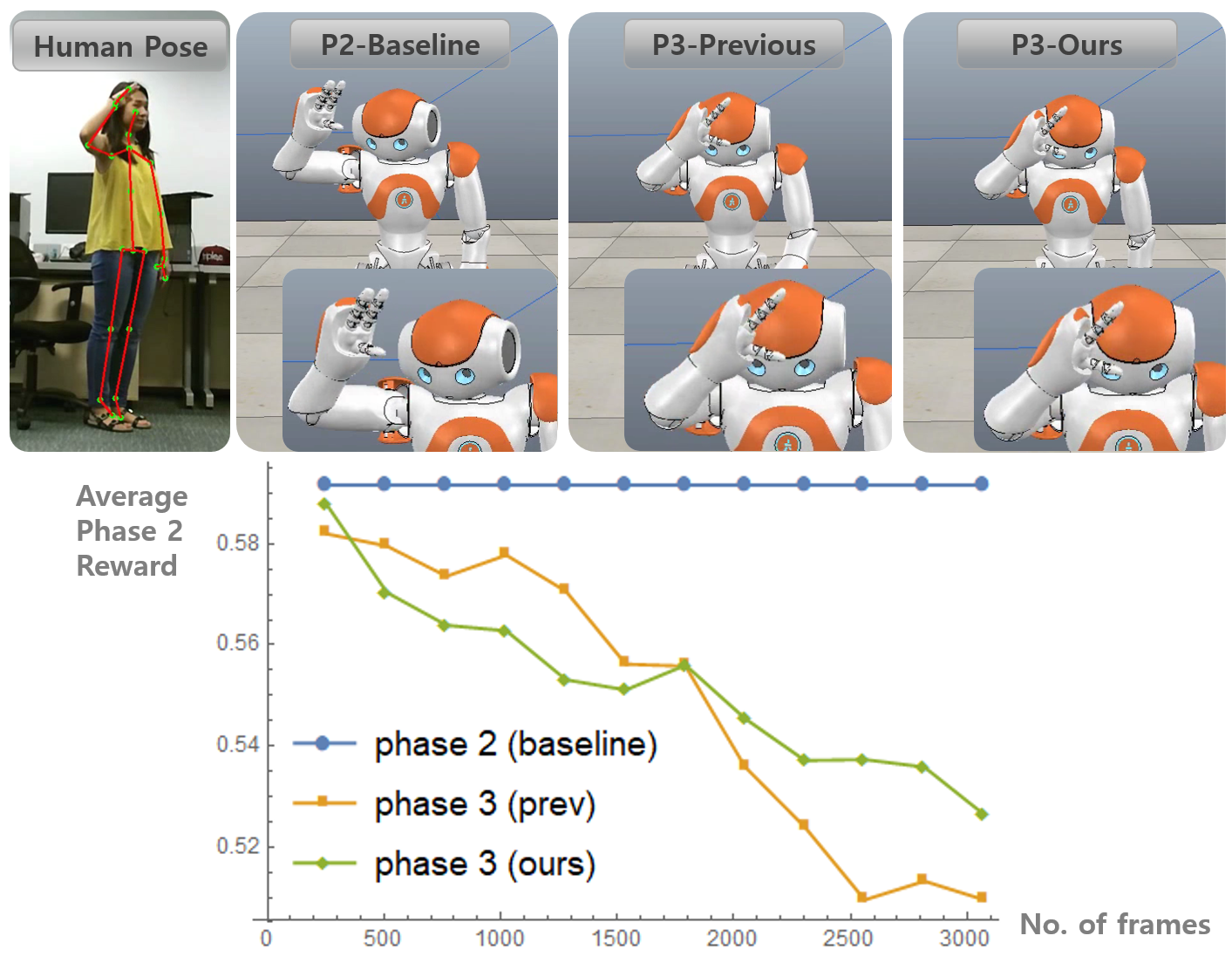}
	\end{center}
	\vspace{-1.0em}
	\caption{DT-based policy fine tuning result. Through our method, the policy was able to learn motion details while retaining much of the retargeting skills learned in phase 2.}
	\vspace{-1.5em}
\end{figure}

\subsection{Policy Optimization Experiments on Phase 3}
In our previous study, policy fine tuning was performed through errors in the joint space. We were able to learn the motion details; however, there was a significant loss of the retargeting skill learned in phase 2. We estimated that the sharp collapse of the policy is caused by the reward space mismatch; \textit{i.e.}, the phase 3 reward is obtained in the joint space while the reward in phase 2 is based on the Cartesian space. To learn the motion details while retaining the learned skill as much as possible, we optimized the policy by fine tuning in the common latent space using a cyclic path. The ground truth datasets of 3k frames for all six motion classes were sampled and shuffled scene-by-scene. Except for the learning rate of the actor ($2.0 \times 10^{-4}$) and the rollout steps (256), the remainder of the other learning parameters were identical to those in phase 2. Based on this experimental environment, we were successfully able to correct our policy as shown in Fig. 7. As the fine tuning progresses, we lose the motion retargeting skill of phase 2. However, learning by cyclic path where the reward is calculated by the distance in the latent space is helpful to keep the motion retargeting skill of phase 2. Compared to our previous work, we achieved great advances in our motion retargeting task; we used a smaller training dataset with the unified encoder-decoder and policy while retaining the pre-learned skill more than our previous work. Qualitative results can be found in our supplementary video: \texttt{https://youtu.be/C37Fip1X0Y0}.

\section{DISCUSSION AND CONCLUSION}
In this study, we proposed the C-3PO method for human-robot motion retargeting. In comparison with the previous work, we achieved a significant improvement in performance through the cyclic and filtering paths, and $n$-step MC method. However, forgetting of previously learned skill still remains unsolved during the direct teaching, and the motion ambiguity problem occasionally occurred due to frame-by-frame approach. In the future work, we will attempt trajectory-based motion retargeting to overcome these problems. We expect that the proposed framework can be extended to object-involved robotic tasks such as pick-and-place. For example, picking object motion can be generated by existing motion retargeting policy and used for ground-truth of initial robot motion learning. Also, from the paired human skeleton data, our approach can be applied to learn motion generation skill for human-robot interaction (HRI) such as handshaking.

\section*{ACKNOWLEDGMENT}

This work was supported both by UST Young Scientist Research Program. (No. [18AS1810]) and by the ICT R\&D program of MSIP/IITP. [2017-0-00162], Development of Human-care Robot Technology for Aging Society.

\bibliographystyle{IEEEtr}
\bibliography{root}

\begin{thebibliography}{10}

\bibitem{shahverdi2016simple}
P.~Shahverdi and M.~T. Masouleh, ``A simple and fast geometric kinematic
  solution for imitation of human arms by a nao humanoid robot,'' in {\em 2016
  4th International Conference on Robotics and Mechatronics (ICROM)},
  pp.~572--577, IEEE, 2016.

\bibitem{zuher2012recognition}
F.~Zuher and R.~Romero, ``Recognition of human motions for imitation and
  control of a humanoid robot,'' in {\em 2012 Brazilian Robotics Symposium and
  Latin American Robotics Symposium}, pp.~190--195, IEEE, 2012.

\bibitem{lee2012full}
J.-H. Lee {\em et~al.}, ``Full-body imitation of human motions with kinect and
  heterogeneous kinematic structure of humanoid robot,'' in {\em 2012 IEEE/SICE
  International Symposium on System Integration (SII)}, pp.~93--98, IEEE, 2012.

\bibitem{monzani2000using}
J.-S. Monzani, P.~Baerlocher, R.~Boulic, and D.~Thalmann, ``Using an
  intermediate skeleton and inverse kinematics for motion retargeting,'' in
  {\em Computer Graphics Forum}, vol.~19, pp.~11--19, Wiley Online Library,
  2000.

\bibitem{mukherjee2015inverse}
S.~Mukherjee, D.~Paramkusam, and S.~K. Dwivedy, ``Inverse kinematics of a nao
  humanoid robot using kinect to track and imitate human motion,'' in {\em 2015
  International Conference on Robotics, Automation, Control and Embedded
  Systems (RACE)}, pp.~1--7, IEEE, 2015.

\bibitem{craig2009introduction}
J.~J. Craig, {\em Introduction to robotics: mechanics and control, 3/E}.
\newblock Pearson Education India, 2009.

\bibitem{lei2015whole}
J.~Lei, M.~Song, Z.-N. Li, and C.~Chen, ``Whole-body humanoid robot imitation
  with pose similarity evaluation,'' {\em Signal Processing}, vol.~108,
  pp.~136--146, 2015.

\bibitem{cole2007learning}
J.~B. Cole, D.~B. Grimes, and R.~P. Rao, ``Learning full-body motions from
  monocular vision: Dynamic imitation in a humanoid robot,'' in {\em 2007
  IEEE/RSJ International Conference on Intelligent Robots and Systems},
  pp.~240--246, IEEE, 2007.

\bibitem{koenemann2014real}
J.~Koenemann, F.~Burget, and M.~Bennewitz, ``Real-time imitation of human
  whole-body motions by humanoids,'' in {\em 2014 IEEE International Conference
  on Robotics and Automation (ICRA)}, pp.~2806--2812, IEEE, 2014.

\bibitem{ott2008motion}
C.~Ott, D.~Lee, and Y.~Nakamura, ``Motion capture based human motion
  recognition and imitation by direct marker control,'' in {\em Humanoids
  2008-8th IEEE-RAS International Conference on Humanoid Robots}, pp.~399--405,
  IEEE, 2008.

\bibitem{yamane2009simultaneous}
K.~Yamane and J.~Hodgins, ``Simultaneous tracking and balancing of humanoid
  robots for imitating human motion capture data,'' in {\em 2009 IEEE/RSJ
  International Conference on Intelligent Robots and Systems}, pp.~2510--2517,
  IEEE, 2009.

\bibitem{kim2009stable}
S.~Kim, C.~Kim, B.~You, and S.~Oh, ``Stable whole-body motion generation for
  humanoid robots to imitate human motions,'' in {\em 2009 IEEE/RSJ
  International Conference on Intelligent Robots and Systems}, pp.~2518--2524,
  IEEE, 2009.

\bibitem{zhang2018deep}
T.~Zhang, Z.~McCarthy, O.~Jowl, D.~Lee, X.~Chen, K.~Goldberg, and P.~Abbeel,
  ``Deep imitation learning for complex manipulation tasks from virtual reality
  teleoperation,'' in {\em 2018 IEEE International Conference on Robotics and
  Automation (ICRA)}, pp.~1--8, IEEE, 2018.

\bibitem{baek2003motion}
S.~Baek, S.~Lee, and G.~J. Kim, ``Motion retargeting and evaluation for
  vr-based training of free motions,'' {\em The Visual Computer}, vol.~19,
  no.~4, pp.~222--242, 2003.

\bibitem{grunwald2003programming}
G.~Grunwald, G.~Schreiber, A.~Albu-Schaffer, and G.~Hirzinger, ``Programming by
  touch: The different way of human-robot interaction,'' {\em IEEE Transactions
  on Industrial Electronics}, vol.~50, no.~4, pp.~659--666, 2003.

\bibitem{kushida2001human}
D.~Kushida, M.~Nakamura, S.~Goto, and N.~Kyura, ``Human direct teaching of
  industrial articulated robot arms based on force-free control,'' {\em
  Artificial Life and Robotics}, vol.~5, no.~1, pp.~26--32, 2001.

\bibitem{tsumugiwa2002variable}
T.~Tsumugiwa, R.~Yokogawa, and K.~Hara, ``Variable impedance control based on
  estimation of human arm stiffness for human-robot cooperative calligraphic
  task,'' in {\em Proceedings 2002 IEEE International Conference on Robotics
  and Automation (Cat. No. 02CH37292)}, vol.~1, pp.~644--650, IEEE, 2002.

\bibitem{schraft2005powermate}
R.~D. Schraft, C.~Meyer, C.~Parlitz, and E.~Helms, ``Powermate-a safe and
  intuitive robot assistant for handling and assembly tasks,'' in {\em
  Proceedings of the 2005 IEEE International Conference on Robotics and
  Automation}, pp.~4074--4079, IEEE, 2005.

\bibitem{kim2019teachme}
T.~Kim and J.-H. Lee, ``Teachme: Three-phase learning framework for robotic
  motion imitation based on interactive teaching and reinforcement learning,''
  in {\em 2019 28th IEEE International Conference on Robot and Human
  Interactive Communication (RO-MAN)}, pp.~1--8, IEEE, 2019.

\bibitem{ghadirzadeh2017deep}
A.~Ghadirzadeh, A.~Maki, D.~Kragic, and M.~Bj{\"o}rkman, ``Deep predictive
  policy training using reinforcement learning,'' in {\em 2017 IEEE/RSJ
  International Conference on Intelligent Robots and Systems (IROS)},
  pp.~2351--2358, IEEE, 2017.

\bibitem{mensink2008characterization}
A.~Mensink, ``Characterization and modeling of a dynamixel servo,'' {\em
  Trabajo Individual de Investigaci{\'o}n en el Electrical Engineering Control
  Engineeringde la University of Twente}, 2008.

\bibitem{bubnov2015iterative}
A.~Bubnov, V.~Emashov, and A.~Chudinov, ``Iterative method of measurement with
  a given accuracy for angular velocity errors,'' in {\em 2015 International
  Siberian Conference on Control and Communications (SIBCON)}, pp.~1--4, IEEE,
  2015.

\bibitem{bandera2012survey}
J.~Bandera, J.~Rodriguez, L.~Molina-Tanco, and A.~Bandera, ``A survey of
  vision-based architectures for robot learning by imitation,'' {\em
  International Journal of Humanoid Robotics}, vol.~9, no.~01, p.~1250006,
  2012.

\bibitem{gleicher1998retargetting}
M.~Gleicher, ``Retargetting motion to new characters,'' in {\em Proceedings of
  the 25th annual conference on Computer graphics and interactive techniques},
  pp.~33--42, ACM, 1998.

\bibitem{choi2000online}
K.-J. Choi and H.-S. Ko, ``Online motion retargetting,'' {\em The Journal of
  Visualization and Computer Animation}, vol.~11, no.~5, pp.~223--235, 2000.

\bibitem{hsieh2005motion}
M.-K. Hsieh, B.-Y. Chen, and M.~Ouhyoung, ``Motion retargeting and transition
  in different articulated figures,'' in {\em Ninth International Conference on
  Computer Aided Design and Computer Graphics (CAD-CG'05)}, pp.~6--pp, IEEE,
  2005.

\bibitem{baran2007automatic}
I.~Baran and J.~Popovi{\'c}, ``Automatic rigging and animation of 3d
  characters,'' in {\em ACM Transactions on graphics (TOG)}, vol.~26, p.~72,
  ACM, 2007.

\bibitem{hecker2008real}
C.~Hecker, B.~Raabe, R.~W. Enslow, J.~DeWeese, J.~Maynard, and K.~van Prooijen,
  ``Real-time motion retargeting to highly varied user-created morphologies,''
  in {\em ACM Transactions on Graphics (TOG)}, vol.~27, p.~27, ACM, 2008.

\bibitem{park2004example}
M.~J. Park and S.~Y. Shin, ``Example-based motion cloning,'' {\em Computer
  Animation and Virtual Worlds}, vol.~15, no.~3-4, pp.~245--257, 2004.

\bibitem{dariush2008online}
B.~Dariush, M.~Gienger, A.~Arumbakkam, C.~Goerick, Y.~Zhu, and K.~Fujimura,
  ``Online and markerless motion retargeting with kinematic constraints,'' in
  {\em 2008 IEEE/RSJ International Conference on Intelligent Robots and
  Systems}, pp.~191--198, IEEE, 2008.

\bibitem{dariush2009online}
B.~Dariush, M.~Gienger, A.~Arumbakkam, Y.~Zhu, B.~Jian, K.~Fujimura, and
  C.~Goerick, ``Online transfer of human motion to humanoids,'' {\em
  International Journal of Humanoid Robotics}, vol.~6, no.~02, pp.~265--289,
  2009.

\bibitem{wang2017generative}
S.~Wang, X.~Zuo, R.~Wang, F.~Cheng, and R.~Yang, ``A generative human-robot
  motion retargeting approach using a single depth sensor,'' in {\em 2017 IEEE
  International Conference on Robotics and Automation (ICRA)}, pp.~5369--5376,
  IEEE, 2017.

\bibitem{ayusawa2017motion}
K.~Ayusawa and E.~Yoshida, ``Motion retargeting for humanoid robots based on
  simultaneous morphing parameter identification and motion optimization,''
  {\em IEEE Transactions on Robotics}, vol.~33, no.~6, pp.~1343--1357, 2017.

\bibitem{vijayan2018using}
A.~E. Vijayan, S.~Alexanderson, J.~Beskow, and I.~Leite, ``Using constrained
  optimization for real-time synchronization of verbal and nonverbal robot
  behavior,'' in {\em 2018 IEEE International Conference on Robotics and
  Automation (ICRA)}, pp.~1955--1961, IEEE, 2018.

\bibitem{penco2018robust}
L.~Penco, B.~Cl{\'e}ment, V.~Modugno, E.~M. Hoffman, G.~Nava, D.~Pucci, N.~G.
  Tsagarakis, J.-B. Mouret, and S.~Ivaldi, ``Robust real-time whole-body motion
  retargeting from human to humanoid,'' in {\em 2018 IEEE-RAS 18th
  International Conference on Humanoid Robots (Humanoids)}, pp.~425--432, IEEE,
  2018.

\bibitem{mnih2013playing}
V.~Mnih, K.~Kavukcuoglu, D.~Silver, A.~Graves, I.~Antonoglou, D.~Wierstra, and
  M.~Riedmiller, ``Playing atari with deep reinforcement learning,'' {\em arXiv
  preprint arXiv:1312.5602}, 2013.

\bibitem{lample2017playing}
G.~Lample and D.~S. Chaplot, ``Playing fps games with deep reinforcement
  learning,'' in {\em Thirty-First AAAI Conference on Artificial Intelligence},
  2017.

\bibitem{silver2016mastering}
D.~Silver, A.~Huang, C.~J. Maddison, A.~Guez, L.~Sifre, G.~Van Den~Driessche,
  J.~Schrittwieser, I.~Antonoglou, V.~Panneershelvam, M.~Lanctot, {\em et~al.},
  ``Mastering the game of go with deep neural networks and tree search,'' {\em
  nature}, vol.~529, no.~7587, p.~484, 2016.

\bibitem{liu2018imitation}
Y.~Liu, A.~Gupta, P.~Abbeel, and S.~Levine, ``Imitation from observation:
  Learning to imitate behaviors from raw video via context translation,'' in
  {\em 2018 IEEE International Conference on Robotics and Automation (ICRA)},
  pp.~1118--1125, IEEE, 2018.

\bibitem{peng2018deepmimic}
X.~B. Peng, P.~Abbeel, S.~Levine, and M.~van~de Panne, ``Deepmimic:
  Example-guided deep reinforcement learning of physics-based character
  skills,'' {\em ACM Transactions on Graphics (TOG)}, vol.~37, no.~4, p.~143,
  2018.

\bibitem{james2017transferring}
S.~James, A.~J. Davison, and E.~Johns, ``Transferring end-to-end visuomotor
  control from simulation to real world for a multi-stage task,'' {\em arXiv
  preprint arXiv:1707.02267}, 2017.

\bibitem{quillen2018deep}
D.~Quillen, E.~Jang, O.~Nachum, C.~Finn, J.~Ibarz, and S.~Levine, ``Deep
  reinforcement learning for vision-based robotic grasping: A simulated
  comparative evaluation of off-policy methods,'' in {\em 2018 IEEE
  International Conference on Robotics and Automation (ICRA)}, pp.~6284--6291,
  IEEE, 2018.

\bibitem{faust2018prm}
A.~Faust, K.~Oslund, O.~Ramirez, A.~Francis, L.~Tapia, M.~Fiser, and
  J.~Davidson, ``Prm-rl: Long-range robotic navigation tasks by combining
  reinforcement learning and sampling-based planning,'' in {\em 2018 IEEE
  International Conference on Robotics and Automation (ICRA)}, pp.~5113--5120,
  IEEE, 2018.

\bibitem{levine2016end}
S.~Levine, C.~Finn, T.~Darrell, and P.~Abbeel, ``End-to-end training of deep
  visuomotor policies,'' {\em The Journal of Machine Learning Research},
  vol.~17, no.~1, pp.~1334--1373, 2016.

\bibitem{schulman2017proximal}
J.~Schulman, F.~Wolski, P.~Dhariwal, A.~Radford, and O.~Klimov, ``Proximal
  policy optimization algorithms,'' {\em arXiv preprint arXiv:1707.06347},
  2017.

\bibitem{kingma2013auto}
D.~P. Kingma and M.~Welling, ``Auto-encoding variational bayes,'' {\em arXiv
  preprint arXiv:1312.6114}, 2013.

\bibitem{konda2000actor}
V.~R. Konda and J.~N. Tsitsiklis, ``Actor-critic algorithms,'' in {\em Advances
  in neural information processing systems}, pp.~1008--1014, 2000.

\bibitem{shahroudy2016ntu}
A.~Shahroudy, J.~Liu, T.-T. Ng, and G.~Wang, ``Ntu rgb+ d: A large scale
  dataset for 3d human activity analysis,'' in {\em Proceedings of the IEEE
  conference on computer vision and pattern recognition}, pp.~1010--1019, 2016.

\bibitem{rohmer2013v}
E.~Rohmer, S.~P. Singh, and M.~Freese, ``V-rep: A versatile and scalable robot
  simulation framework,'' in {\em 2013 IEEE/RSJ International Conference on
  Intelligent Robots and Systems}, pp.~1321--1326, IEEE, 2013.

\bibitem{pot2009choregraphe}
E.~Pot, J.~Monceaux, R.~Gelin, and B.~Maisonnier, ``Choregraphe: a graphical
  tool for humanoid robot programming,'' in {\em RO-MAN 2009-The 18th IEEE
  International Symposium on Robot and Human Interactive Communication},
  pp.~46--51, IEEE, 2009.

\bibitem{sutton2018reinforcement}
R.~S. Sutton and A.~G. Barto, {\em Reinforcement learning: An introduction}.
\newblock MIT press, 2018.

\end{thebibliography}

\end{document}